%% file: arxiv.tex
\documentclass[10pt,twocolumn,letterpaper]{article}

\usepackage{iccv}              


\definecolor{iccvblue}{rgb}{0.21,0.49,0.74}
\usepackage[pagebackref,breaklinks,colorlinks,allcolors=iccvblue]{hyperref}
\usepackage{graphicx}

\usepackage[accsupp]{axessibility}  

\usepackage{multirow}
\usepackage{caption}
\def\paperID{8429} 
\def\confName{ICCV}
\def\confYear{2025}

\title{D3: Training-Free AI-Generated Video Detection Using Second-Order Features}
\author{
    \textbf{Chende Zheng}$^{1}$ \quad 
    \textbf{Ruiqi Suo}$^{1}$ \quad 
    \textbf{Chenhao Lin}$^{1}$\thanks{Corresponding Authors} \quad
    \textbf{Zhengyu Zhao}$^{1}$ \quad 
    \textbf{Le Yang}$^{1}$ \quad \\
    \textbf{Shuai Liu}$^{1}$\footnotemark[1] \quad 
    \textbf{Minghui Yang}$^{2}$ \quad 
    \textbf{Cong Wang}$^{3}$ \quad 
    \textbf{Chao Shen}$^{1}$ \\
    $^{1}$Xi’an Jiaotong University \quad $^{2}$Guangdong OPPO Mobile Communications Co., Ltd. \\ 
    $^{3}$City University of Hong Kong
}

\begin{document}

\maketitle

\begin{abstract}\label{abstract}
The evolution of video generation techniques, such as Sora, has made it increasingly easy to produce high-fidelity AI-generated videos, raising public concern over the dissemination of synthetic content. 
However, existing detection methodologies remain limited by their insufficient exploration of temporal artifacts in synthetic videos. 
To bridge this gap, we establish a theoretical framework through second-order dynamical analysis under Newtonian mechanics, subsequently extending the Second-order Central Difference features tailored for temporal artifact detection. Building on this theoretical foundation, we reveal a fundamental divergence in second-order feature distributions between real and AI-generated videos.
Concretely, we propose Detection by Difference of Differences (D3), a novel training-free detection method that leverages the above second-order temporal discrepancies.
We validate the superiority of our D3 on 4 open-source datasets (Gen-Video, VideoPhy, EvalCrafter, VidProM), 40 subsets in total.
For example, on GenVideo, D3 outperforms the previous state-of-the-art method by 10.39\% (absolute) mean Average Precision.
Additional experiments on time cost and post-processing operations demonstrate D3's exceptional computational efficiency and strong robust performance.
Our code is available at \href{https://github.com/Zig-HS/D3}{\textit{https://github.com/Zig-HS/D3}}.
\end{abstract}

\begin{figure}[!t]
    \centering
    \includegraphics[width=0.5\textwidth]{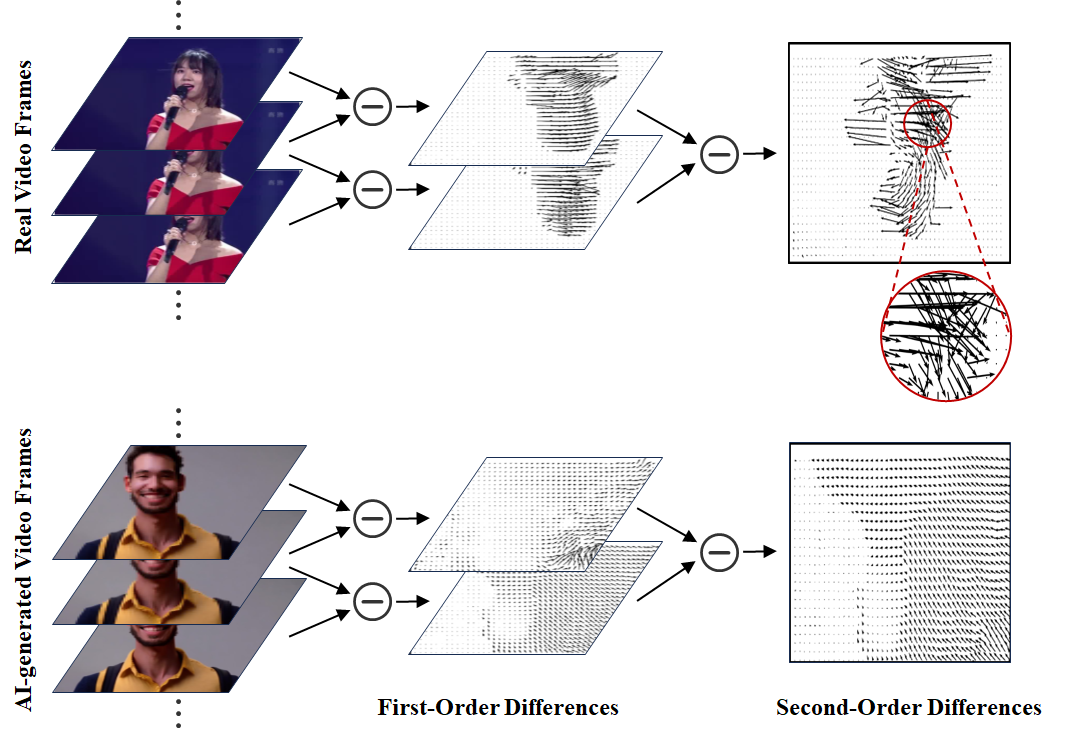}
    \caption{
        Real and AI-generated videos can be differentiated by second-order (temporal) features in our \textbf{D}etection by \textbf{D}ifference of \textbf{D}ifferences (D3) method. The optical flow vectors are obtained by RAFT~\cite{teed2020raft}. First-order features are less powerful (see Table~\ref{tab:order}.) 
    }
    \setlength{\belowcaptionskip}{0pt} 
    \label{fig:flow}
\end{figure}



\section{Introduction}\label{Introduction}
With the development of AI~\cite{zha2025enable,pu2025large,gu2025mr}, generative models have evolved into versatile tools for high-fidelity content creation. However, their pervasive deployment across digital ecosystems has precipitated critical societal security risks—systematically undermining information integrity and eroding public trust~\cite{sharma2023survey,barrett2023identifying,3D2Fool} - evidenced by cases like the Taylor Swift deepfakes~\cite{bbcTaylorSwift}. Consequently, there is an increasing and urgent demand to develop a detector of AI-generated videos.
Traditional research on deep forgery detection focuses on facial forgery (e.g., Deepfakes) but is unable to generalize to universal videos. For this, there has been some recent aiming at the generalization detection of AI-generated videos, e.g., DuB3D~\cite{ji2024distinguish}, DeCoF~\cite{ma2024decof}, DeMamba~\cite{chen2024demamba}, etc. These methods commonly learn the differences between real and AI-generated videos from the training data by using deep learning frameworks.

However, despite these efforts, there is still a lack of interpretability research on AI-generated video detection. To address this, we introduce the second-order position control system under Newtonian mechanics theory. We extend the \textit{Second-order Central Difference} features from this system to AI-generated video detection and validate significant differences in second-order features between real and AI-generated videos by conducting visualization experiments on differential optical flow.
More concretely, we propose Detection by Difference of Differences (D3), a training-free AI-generated video detection framework based on second-order features. 

Technically, we use a pre-trained visual encoder to extract zero-order features from the input video by frames. 
Then, we employ L2 distance (or cosine similarity) to assess the inter-frame differences as first-order features. 
We further compute the second-order features according to the \textit{Second-order Central Difference} formula. 
Our empirical findings show that the second-order features of real videos exhibit more significant volatility, while AI-generated videos tend to follow a flatter pattern (Figure~\ref{fig:flow}). Therefore, we quantify the volatility by calculating the standard deviation of the second-order features and realizing the generalized detection of AI-generated videos. Besides, to comprehensively evaluate the generalization ability of D3, we conduct an open-world evaluation, which validate D3's superior detection performance.


Our main contributions can be summarized as follows:
\begin{itemize}
    \item By rethinking Newtonian mechanics, we innovatively introduce the second-order central difference features of videos. Our experiments reveal the differences in second-order features between existing AI-generated videos and real videos.
    \item We propose D3, a novel training-free AI-generated video detection method. By extracting second-order features, our detector is capable of generalizing across various generators.
    \item 
    The extensive experimental results on 4 different open-source datasets, including 40 test subsets, demonstrate the state-of-the-art (SOTA) generalization performance as well as strong robustness to the post-processing operations of our method.
\end{itemize}

\section{Related Work}\label{Related Work}

\subsection{Video Generation Methods} 

Recent advances in video synthesis, predominantly built on diffusion models, focus on text-to-video, image-to-video, and combined text-image approaches.
A core challenge remains ensuring logical coherence and smooth temporal continuity. To address this, Text2Video-Zero~\cite{Khachatryan_2023_ICCV} enriches latent codes with motion dynamics, while Zhang et al.~\cite{zhang2023i2vgen} leverage visual guidance in I2VGen-XL for coherent high-resolution generation. Similarly, Xing et al.~\cite{xing2025dynamicrafter} and Chen et al.~\cite{chen2023videocrafter1} incorporate motion cues—via video diffusion priors and generation-stage cues, respectively—to simulate realistic motion and enhance frame transitions. For long videos, SEINE~\cite{chen2023seine} automates smooth transitions using scene images with text control.

Regarding datasets, SVD~\cite{blattmann2023stable} demonstrates that fine-tuning on small, curated datasets yields higher-quality, stable models compared to non-curated alternatives.

While these video generation methods have yielded promising results, we have found that they still produce videos that do not fully comply with the physical laws of the real world. Therefore, there is still significant room for improvement in video generation.

\subsection{AI-Generated Video Detection}

As video synthesis quality advances, effective AIGC detection for videos becomes increasingly critical. 
Traditional deepfake detection focuses on facial artifacts (e.g., distortions in landmarks~\cite{yang2019exposing} or head pose inconsistencies~\cite{li1811exposing}). These methods struggle with complex scenes. Recent approaches shift toward global characteristics: NPR~\cite{tan2024rethinking} analyzes neighboring pixel relationships, while others leverage pre-trained models~\cite{aghasanli2023interpretable,ojha2023towards,tan2023learning} or data augmentation~\cite{jeong2022frepgan,epstein2023online} to improve diffusion-image detection. 

For videos generated by Diffsion Models, recent advances now extend to universal detection, exemplified by DeMamba~\cite{chen2024demamba}, which introduced a dedicated Mamba module for video detection and developed the Gen-video dataset, which was specifically designed for AI-generated video detection tasks. 
In a similar vein, Liu et al.\cite{liu2024turns} proposed a CNN+LSTM architecture that utilizes DIRE\cite{10377654} residuals to classify videos as real or generated, while DeCoF~\cite{ma2024decof} presented a detector that focuses on temporal artifacts, aiming to identify AI-generated videos by analyzing these specific features. 
These approaches contribute to the growing field of AI-generated video detection, each addressing unique aspects of the challenge.

However, although these methods strive to distinguish between real and AI-generated videos, they still lack a deep analysis of temporal artifacts, resulting in a missing enlightening motivation for the interpretable detection of AI-generated videos.

\section{Methodology}\label{Methodology}

\subsection{Motivation}\label{Motivation}

\begin{figure*}[!t]
    \centering
    \includegraphics[width=1\textwidth]{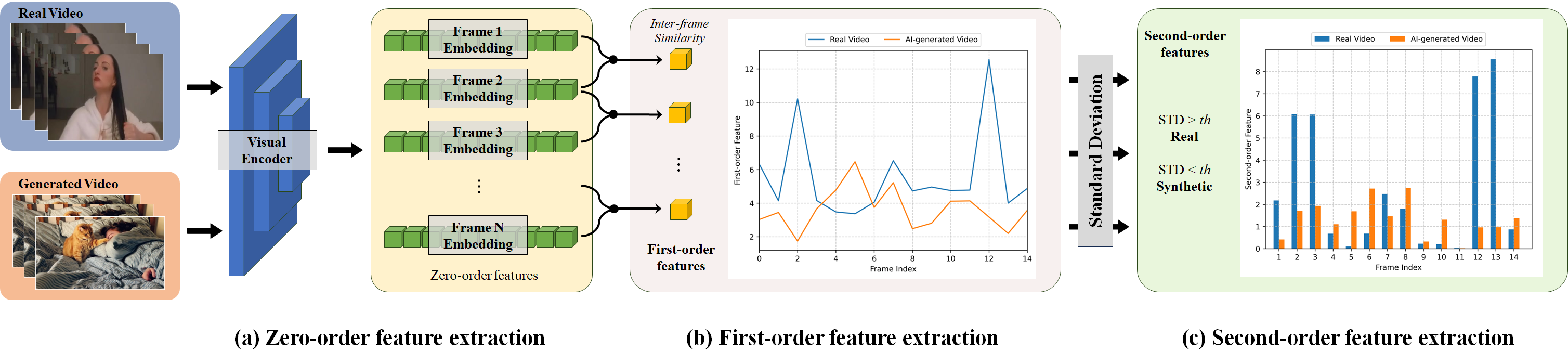}
        \caption{Framework of our training-free detection method D3. For a given video, (a) zero-order, (b) first-order, and (c) second-order features are subsequently extracted.}
    \label{fig:pipeline}
\end{figure*}


The key to detecting AI-generated videos lies in identifying the artifacts that differ from real videos. Existing detection methods focus on pixel-level temporal artifacts of specific regions (e.g., lips and facial edges). However, as the quality of the generated video improves, the generalization performance of these methods continually decreases. 
An effective solution is to analyze the artifacts from a theoretical perspective. For this, we propose an analysis method based on second-order features to investigate the differences in second-order features between real and AI-generated videos. Specifically, we start by modeling a second-order position control system under Newtonian mechanics in the real world, which can be represented by the following equation:
\begin{align}
    A_2\frac{d^2x(t)}{dt^2} + A_1\frac{dx(t)}{dt} + A_0x(t) = u(t)
\end{align}
where $A_2$ is the inertia coefficient (i.e. second-order coefficient), $A_1$ is the damping coefficient, $A_0$ is the elasticity coefficient, and $u(t)$ represents the input force.
Note that real-world systems are typically higher-order systems, but they can be reduced to second-order systems using the \textit{Dominant Pole Approximation}~\cite{ogata2020modern}).

According to \textit{the Principles of Automatic Control}, when solving a second-order control system, we use the second-order central difference method to approximate the derivative of the differential equation. In other words, the second-order ordinary differential equation can be discretized to obtain a numerical solution. Therefore, we can approximate the acceleration $f''(x)$ (i.e. second-order feature) using the \textit{Second-order Central Difference}~\cite{ogata2020modern} as follows:
\begin{align}\label{formula:2d}
    f''(x)&=\frac{f(x+h)-2f(x)+f(x-h)}{h^2} \\
&=\frac{f'(x)-f'(x-h)}{h}
\end{align}
where $x$ is the time point, $h$ is the sampling interval, and $f'(x)$ represents the first-order difference feature.

A correct mechanical simulation model should adhere to the physical laws of the real world, which means that the simulated acceleration (i.e., the second-order feature) should follow the same paradigm as real-world objects. 
Therefore, a video generation model is expected to ensure the second-order features of the videos to exhibit similar patterns in real videos.

To verify whether existing video generators can accurately fit the second-order features of real videos, we conduct a visualization experiment on both real and AI-generated videos by extracting optical flow, using RAFT~\cite{teed2020raft}. In this experiment, we extract two optical flow features from the video $ X = {x_1, x_2, \dots, x_N} $. The first optical flow is calculated between frames $ x_t $ and $ x_{t+1} $, while the second optical flow is calculated between frames $ x_{t+1} $ and $ x_{t+2} $. These optical flows represent the speed of pixel change over time, i.e., the first-order difference feature in Formula (\ref{formula:2d}). Therefore, the difference between these two optical flows, $ X_{diff} $, reflects the change in optical flow speed from frame $ x_t $ to frame $ x_{t+2} $ (i.e., optical flow acceleration), and is expressed by the following formula:
\begin{align}
    X_{diff}=\frac{OF(x_{t+1},x_{t+2})-OF(x_t,x_{t+1})}{{\Delta t}^2}
\end{align}
where $ OF(x, y) $ represents the optical flow between frames $ x $ and $ y $, and $ \Delta t $ is the sampling interval.

Figure~\ref{fig:flow} displays the visualized optical flow and the corresponding vector diagrams. The results reveal a clear distinction in the second-order features between real videos and AI-generated videos. 
Real videos exhibit more chaotic speed variations, as shown in Figure~\ref{fig:flow} (1)-(d) \& (2)-(d), reflecting the higher complexity and diversity driven by various influencing factors in real-world scenes. 
In contrast, generated videos tend to show very flat patterns, which could be due to existing generators' difficulty in simulating second-order dynamics, resulting in "smoother" video outputs constrained within the distribution range of their training data. 
These observations support our hypothesis that current generators fail to accurately replicate real videos' second-order features, providing motivation and opportunity for AI-generated video detection through second-order feature analysis.

\subsection{Detection by Difference of Differences}
We have confirmed the distortion of second-order features in existing video generators on pixel level and optical flow level, however, these features are challenging to compute directly. A feasible approach is to use a visual encoder to perform feature dimensionality reduction on video frames, transforming pixel-level features into deep representations. 
Based on this, we introduce a training-free mathematical detection method based on second-order features, aiming to realize the Detection on Difference of Differences (D3).

Given an input video $ X\in \mathbb{R}^{T\times 3\times H\times W} $, which is sampled into a sequence of frames $X = {X_1, X_2, \ldots, X_T}$ at a regular intervals of $\Delta t$. We utilize visual encoders (e.g., DINOv2, XCLIP or pre-trained ResNet-18, etc.) to encode the input frames into a sequence of features $F_0 = {F_0^1, \ldots, F_0^T}, F_0 \in \mathbb{R}^{T \times N}$. Within the feature space, the first-order features are first extracted. Specifically, we use \textit{L2 Distance} or \textit{Cosine Similarity} to calculate inter-frame similarity, as the first-order difference features, formulated as follows:
\begin{align}
F_1^{L2}(k) &= \frac{{dis(F_0^k,F_0^{k+1})}}{\Delta t}, \quad k = 1, 2, ..., T-1 \\
F_1^{Cos}(k) &= \frac{{sim(F_0^k,F_0^{k+1})}}{\Delta t}, \quad k = 1, 2, ..., T-1
\end{align}
where $ k $ is the frame index.
Then, we further compute the second-order central difference feature according to Formula (\ref{formula:2d}), as follows:
\begin{align}
F_2(k) = \frac{F_1(k)-F_1(k-1)}{\Delta t}, \quad k = 2, ..., T-1
\end{align}

We present the first-order and second-order features (absolute values) extracted from real and AI-generated videos in Figure~\ref{fig:pipeline}. The results show that, compared to generated videos, the temporal second-order features of real videos exhibit more pronounced fluctuations, which is consistent with the conclusion drawn in Figure~\ref{fig:flow}.

To measure this volatility, we calculate the standard deviation of the second-order features, as the following formula:
\begin{align}
\sigma(F_2) = \sqrt{\frac{1}{T-3} \sum_{i=2}^{T-1} (F_2(i) - \frac{1}{T-3} \sum_{i=2}^{T-1} F_2(i))^2}
\end{align}

We use this standard deviation as the final output to classify real and generated videos. The overall pipeline of D3 is shown in Figure~\ref{fig:pipeline}. Compared to existing image or video detection methods, our approach has significant advantages: A) D3 is training-free, consisting solely of an inference process, and does not require generated videos. B) D3 demonstrates exceptional computational efficiency, with the primary computational cost stemming from visual feature extraction. Furthermore, our experimental results (see Section~\ref{Effect of Visual Encoders}) indicate that D3 remains effective even when using lightweight feature extractors.

\begin{table*}[!t]
    \centering
    \resizebox{1\linewidth}{!}{
        \begin{tabular}{lccccccccccccc}
            \toprule
            Detection& Detection & \multicolumn{10}{c}{Datasets (AP$\uparrow$)}  & \multirow{2}{*}{mAP}\\ 
            \cmidrule(r){3-12}
            Method & Level & Crafter& Gen2 & HotShot & Lavie & MSE & MV & MSO & Show-1 & Sora & WS \\
            \midrule 
            FID~\cite{zheng2025breaking} & Image & 92.41 & 93.27 & \underline{86.10} & 83.68 & \underline{91.50} & 93.67 & 92.24 & \underline{90.61} & 74.95 & \underline{82.24} & \underline{88.07}\\
            NPR~\cite{tan2023rethinking} & Image & 97.02  & 96.35  & 40.17  & 22.37  & 84.67  & 96.79  & 96.53  & 21.61  & \underline{90.55}  & 66.51  & 71.26\\
            STIL~\cite{gu2021spatiotemporal} & Image & 85.82  & 93.19  & 40.61  & 53.24  & 58.99  & 94.94  & 71.62  & 47.73  & 22.35  & 61.91  & 63.04 \\ 
            MINITIME~\cite{coccomini2024mintime} & Video & 88.62  & 60.66  & 39.03  & 82.29  & 23.85  & 74.79  & 74.33  & 41.08  & 16.92  & 72.25  & 57.38 \\
            FTCN~\cite{zheng2021exploring} & Video & 95.41  & 97.18  & 37.47  & 44.90  & 79.71  & 99.75  & 97.05  & 17.33  & 83.69  & 66.86  & 71.94\\
            TALL~\cite{xu2023tall} & Video & 87.85  & 93.47  & 44.00  & 59.07  & 51.11  & 92.09  & 63.63  & 51.06  & 15.82  & 64.43  & 62.25\\
            XCLIP~\cite{ni2022expanding} & Video & {97.32}  & \textbf{99.44}  & 44.68  & 72.69  & 88.00  & \textbf{99.96}  & 97.53  & 38.37  & 71.08  & 74.00  & 78.31\\
            AIGVDet~\cite{bai2024ai} & Video & 75.87  & 89.98  & {51.81}  & \underline{88.62}  & 70.91  & 56.22  & 67.93  & 72.59  & 65.70  & 64.96  & 70.46\\
            Demamba~\cite{chen2024demamba} & Video & \underline{97.91} & {99.16}  & {52.97}  & 76.72  & 82.83  & \underline{99.80}  & \underline{98.42}  & 56.24 & 77.75 & 74.81 & 81.66\\
            Our D3 & Video & \textbf{98.53}  & \underline{99.39}  & \textbf{98.52}  & \textbf{97.22}  & \textbf{97.12}  & 99.52  & \textbf{98.68}  & \textbf{99.18}  & \textbf{99.91}  & \textbf{96.49}  & \textbf{98.46} \\
            \bottomrule
        \end{tabular}
    }
    \caption{
        Detection results on GenVideo datasets. 
        Our D3 is training-free, while the baselines are trained on real videos from Youku-mPLUG~\cite{xu2023youku} and AI-generated videos from Pika~\cite{pikaPika}, following the setting in Demamba~\cite{chen2024demamba}. \textbf{Bold} represents the best and \underline{underline} represents the second best.
    }
    \label{tab:genvideo}
\end{table*}

\section{Experiments}\label{Experiments}
\subsection{Experimental Setup}
\paragraph{Training Datasets.} 

Considering that D3 operates solely during inference, the experiments require no training datasets. However, for a comprehensive comparison, we still set up a training dataset for baselines. Specifically, following the settings from DeMamba~\cite{chen2024demamba}, we use real videos from Youku-mPLUG~\cite{xu2023youku} and AI-generated videos from Pika~\cite{pikaPika} to train baselines.

\paragraph{Test Datasets.} 
To assess the generalization ability of our approach to real-world scenarios, we adopt the 4 out-of-distribution datasets, including 40 test sets: 

\begin{itemize}

    \item \textbf{GenVideo}~\cite{chen2024demamba}: ModelScope (MSE)~\cite{wang2023modelscope}, MorphStudio (MSO)~\cite{morphstudioAllinoneVideo}, MoonValley (MV)~\cite{moonvalleyMoonvalley}, HotShot~\cite{huggingfaceHotshotcoHotshotXLHugging}, Show\_1~\cite{zhang2024show}, Gen2~\cite{esser2023structure}, Crafter~\cite{chen2023videocrafter1}, LaVie~\cite{wang2023lavie}, Sora~\cite{brooks2024video}, and WildScrape (WS)~\cite{wei2024dreamvideo}. 
    
    \item \textbf{EvalCrafter}~\cite{liu2024evalcrafter}: MoonValley (MV), Floor33~\cite{Floor33}, Gen2, Gen2-December (Gen2-Dec), HotShot, LaVie-Base (LaVie-B), LaVie-Internet (LaVie-I), Mix-SR, ModelScope, Pika, Pika\_v1, Show\_1, VideoCrafter (VC)~\cite{chen2023videocrafter1}, and ZeroScope (ZS)~\cite{huggingfaceCerspensezeroscope_v2_576wHugging}.

    \item \textbf{VideoPhy}~\cite{bansal2024videophy}: LaVie, OpenSora\cite{opensora}, CogVideoX\cite{yang2024cogvideox}, CogVideoX-5B, Dream-Machine\cite{dream-machine}, Gen2, Pika, SVD\cite{blattmann2023stable}, VideoCrafter2 (VC2)\cite{chen2024videocrafter2}, and ZeroScope.
    
    \item \textbf{VidProM}~\cite{wang2024vidprom}: ModelScope (MSE), OpenSora (OS), Pika, StreamingT2V (ST2V)~\cite{henschel2024streamingt2v}, Text2video-zero (T2VZ)\cite{khachatryan2023text2video}, and VideoCrafter2 (VC2).
    
\end{itemize}

\paragraph{Baselines.} 
We perform comparisons of our approach with existing popular and state-of-the-art detectors, including 
3 image-level detectors, FID (NeurIPS'24)~\cite{zheng2025breaking}, NPR (CVPR'24)~\cite{tan2023rethinking} and STIL (MM'21)~\cite{gu2021spatiotemporal}, 
6 video-level detectors, FTCN (ICCV'21)~\cite{zheng2021exploring}, MINITIME (TIFS'24)~\cite{coccomini2024mintime}, TALL (ICCV'23)~\cite{xu2023tall}, XCLIP (ECCV'22)~\cite{ni2022expanding}, AIGVDet~\cite{bai2024ai}, and DeMamba~\cite{chen2024demamba}.

We re-implement baselines~\cite{zheng2025breaking,tan2023rethinking,bai2024ai,chen2024demamba} with the official codes using our training set.
We report the results of the baselines~\cite{gu2021spatiotemporal,zheng2021exploring,coccomini2024mintime,xu2023tall,ni2022expanding,chen2024demamba} on GenVideo dataset from~\cite{chen2024demamba}.

\paragraph{Implementation Details.}

We adopt pre-trained XCLIP-B/16~\cite{ni2022expanding} as the visual encoders to extract zero-order features and L2 Distance to calculate first-order features. 
During inference, 
we extract a segment from the input video (up to 2 seconds) and frames are sampled at equal intervals of 8 frames per second. 
All frames are set to JPEG format. For pre-processing, we crop 10\% of the longer edge of all frames and then resize the frames to $224\times 224$ pixels. 
We adopt the Average Precision (AP) and the Area Under the Receiver Operating Characteristic curve (AUROC, AUC) as the evaluation metric, which is widely used in baselines~\cite{tan2023rethinking,chen2024demamba,gu2021spatiotemporal,zheng2021exploring,coccomini2024mintime,xu2023tall,ni2022expanding}. (AUC results are provided in the Supplementary Materials.)
All of our experiments are conducted using PyTorch on AMD EPYC 7763 64-Core CPU and NVIDIA GeForce RTX 4090 Tensor Core GPU. 

\begin{table*}[!t]
\Huge
    \centering
    \resizebox{1\linewidth}{!}{
        \begin{tabular}{lcccccccccccccccc}
            \toprule
           Detection  & \multicolumn{14}{c}{Datasets (AP$\uparrow$)}  & \multirow{2}{*}{mAP}\\ 
            \cmidrule(r){2-15}
            Method & MV & Floor32 & Gen2 & Gen2-D & HotShot & LaVie-V & LaVie-I & Mix-SR & MSE & Pika & Pika-v1 & Show-1 & VC & ZS \\
            \midrule 
            FID & 98.29 & 96.4 & 97.36 & 98.68 & \underline{89.9} & \underline{92.92} & \underline{84.19} & 98.51 & 95.74 & \underline{99.49} & 99.17 & \underline{96.77} & 95.71 & 95.18 & \underline{95.59}  \\  
            NPR & \textbf{99.96}  & \textbf{99.77}  & \underline{99.34}  & \textbf{99.95}  & 47.39  & 76.45  & 72.23  & \textbf{99.67}  & \textbf{98.54}  & \textbf{99.97}  & \textbf{99.93}  & 69.82  & \textbf{99.68}  & \underline{98.21}  & 90.07 \\  
            AIGVDet & 56.50  & 67.84  & 71.86  & 74.24  & 51.46  & 73.81  & 70.72 & 57.64  & 71.00  & 94.95  & 92.92  & 72.41  & 64.58  & 67.00 & 70.50 \\  
            Demamba & 99.49  & 91.76  & 96.98  & 99.27  & 34.60  & 56.89  & 37.85  & 97.49  & 71.33  & 98.69  & 99.33  & 26.83  & 94.30  & 64.39  & 76.37 \\ 
            Our D3 & \underline{99.52}  & \underline{98.68}  & \textbf{99.46}  & \underline{99.74}  & \textbf{98.52}  & \textbf{97.79}  & \textbf{98.48}  & \underline{99.16}  & \underline{97.13}  & 99.43  & \underline{99.55}  & \textbf{99.18}  & \underline{98.77}  & \textbf{98.83}  & \textbf{98.87} \\ 
            \bottomrule
        \end{tabular}
    }
    \caption{
    Detection results on 14 EvalCrafter datasets.
    }
    \label{tab:evalcrafter}
\end{table*}

\begin{table*}[!t]
    \centering
    \resizebox{1\linewidth}{!}{
        \begin{tabular}{lccccccccccccc}
            \toprule
            Detection & \multicolumn{10}{c}{Datasets (AP$\uparrow$)}  & \multirow{2}{*}{mAP}\\ 
            \cmidrule(r){2-11}
            Method  & LaVie & OpenSora & CogVideoX-5B & CogVideoX & Dream-Machine & Gen-2 & Pika & SVD & VC2 & ZeroScope \\
            \midrule 
           FID  & \underline{96.51} & 87.9 & \underline{91.41} & \underline{93.34} & 97.5 & 98.35 & 99.55 & 95.66 & \underline{96.03} & \underline{90.6} & \underline{94.69} \\ 
           NPR  & 63.72  & \underline{88.78}  & 81.99  & 81.37  & \textbf{99.86}  & \textbf{99.90}  & \textbf{99.91}  & \textbf{99.54}  & 60.21  & 78.23  & 85.35 \\ 
           AIGVDet  & 61.06  & 59.07  & 58.95  & 63.15  & 59.27  & 61.55  & 92.96  & 53.73  & 58.22  & 63.11  & 63.11 \\
           Demamba   & 28.80  & 16.00  & 24.35  & 22.97  & 94.03  & 97.52  & 96.75  & 87.28  & 23.86  & 23.17  & 51.47 \\
           Our D3  & \textbf{98.49}  & \textbf{98.55}  & \textbf{99.03}  & \textbf{98.87}  & \underline{99.54}  & \underline{99.87}  & \underline{99.70}  & \underline{98.75}  & \textbf{99.46}  & \textbf{99.38}  & \textbf{99.16} \\ 
            \bottomrule
        \end{tabular}
    }
    \caption{
        Detection results on 10 VideoPhy datasets.
    }
    \label{tab:videophy}
\end{table*}

\begin{table}[!t]
    \centering
    \resizebox{1\linewidth}{!}{
        \begin{tabular}{lcccccccccccccc}
            \toprule
            Detection & \multicolumn{6}{c}{Datasets (AP$\uparrow$)}  & \multirow{2}{*}{mAP}\\ 
            \cmidrule(r){2-7}
            Method & MSE & OS & Pika & ST2V & T2VZ & VC2 \\
            \midrule 
           FID & \underline{91.35} & 87.68 & \underline{99.59} & \textbf{97.87} & 68.51 & \underline{85.92} & \textbf{88.49}\\
           NPR & 87.04  & \underline{89.85}  & \textbf{99.98}  & 89.88  & \textbf{88.93}  & 70.79  & 87.75\\
           AIGVDet & 63.33  & 62.12  & 66.07  & 55.46  & 63.49  & 52.15  & 60.44\\ 
           Demamba & 58.73  & 85.87  & 99.34  & 86.48  & \underline{79.62}  & 80.28  & 81.72\\
           Our D3 & \textbf{96.85}  & \textbf{97.85}  & 99.14  & \underline{93.13}  & 45.11  & \textbf{98.70}  & \underline{88.46}\\
            \bottomrule
        \end{tabular}
    }
    \caption{
     Detection results on 6 VidProM datasets.
    }
    \label{tab:vidprom}
\end{table}

\subsection{Detection on GenVideo}

We conduct a comprehensive comparative analysis of various AI-generated video detectors, evaluating their generalization performance on the GenVideo dataset. The AP results are presented in Table~\ref{tab:genvideo}. As can be seen, our D3 achieves the base overall results.
In addition, except for FID~\cite{zheng2025breaking}, image-level detectors suffer from performance degradation on generated videos, which is attributed to their inability on video-level temporal artifacts. Meanwhile, MINITIME, FTCN, and TALL also perform poorly on GenVideo, because they are designed for detecting forged facial videos. 

Interestingly, FID demonstrates strong generalization, which can be attributed to its unique generalization design. FID focuses on the local feature information of images, which enables it to remain unaffected by specific semantic scenes.
Nevertheless, compared to the latest video detection methods or fine-tuned large-scale visual models, D3 demonstrates superior performance on GenVideo. 
Specifically, the mean AP of the D3 method reached 98.46\%, outperforming state-of-the-art FID by 10.39\% (absolute) mean AP.

Note that D3 is entirely training-free and does not require additional generated videos. The results validate our hypothesis that existing video generators \textbf{cannot} accurately model the second-order features of real videos. Furthermore, based on this hypothesis, we can realize accurate detection by calculating the second-order features using mathematical methods.

\begin{table}[!t]
    \centering
    \resizebox{\columnwidth}{!}{%
    \begin{tabular}{l c c c c c}
        \toprule
        Detection  &  \multicolumn{3}{c}{Time (s,$\downarrow$)} & mAP$\uparrow$\\
        \cmidrule(lr){2-4} \cmidrule(lr){5-5}
        Method& Preprocess & Train & Inference & on GenVideo \\
        \midrule
        FID & Free & 415 & 213 & 88.07\\
        NPR & Free & 256 & 188 & 71.26\\
        AIGVDet & 500 & 642 & 74 & 70.46\\
        Demamba & Free & 196 & 91 & 81.66\\
        \midrule
        D3 (XCLIP-B/16) & Free & Free & \underline{56} & \textbf{98.46}\\
        D3 (MobileNet-v3) & Free & Free & \textbf{40} & \underline{95.47}\\
        \bottomrule
    \end{tabular}
    }
    \caption{Efficiency results on GenVideo with 1000 video samples and batch size of 1. The preprocessing overhead of AIGVDet comes from the optical flow extraction using RAFT.
    For image-level methods (FID, NPR), 8 images form a video.}
    \label{tab:time}
\end{table}

\subsection{Detection on More Challenging Datasets}
%
To further assess the generalizability of D3, we extend the evaluation to three more challenging open-source datasets (EvalCrafter, VideoPhy, VidProM), with results presented in Table~\ref{tab:evalcrafter},~\ref{tab:videophy}, \&~\ref{tab:vidprom}. These results demonstrate the consistent superiority of D3.

In these additional experiments, we can observe that image-level detectors (FID and NPR) exhibit stronger generalization than video-level baselines. This is interesting, as it contradicts conclusions from recent research (e.g., DeCoF~\cite{ma2024decof} and DeMamba~\cite{chen2024demamba}). We attribute this to two factors: 1) FID and NPR are designed for generalization in cross-scene and cross-generator settings; 
2) Due to changes in generative models, AI-generated videos on these datasets exhibit more diverse video artifact features. However, universal image-level artifacts still persist (e.g., upsampling artifacts~\cite{gong2019autogan}).

Besides, we find that D3 performs poorly on the T2VZ dataset. We attribute this to the low generation quality of T2VZ, which leads to poor semantic consistency in the generated videos, making them resemble chaotic images rather than dynamic videos. A detailed discussion is provided in the Supplementary Materials.


Despite this, D3 achieves impressive mean APs, outperforming the best-performing baselines. In sum, the results across 40 test subsets underscore the remarkable generalization capability of D3, which can be attributed to D3's successful identification of the substantial differences in second-order features between real and AI-generated videos, further highlighting the limitations of current video generators in fitting second-order features.

\subsection{Efficiency Comparisons}
In Table~\ref{tab:time}, we present the time costs per 1,000 video samples for baselines and D3 across different stages. The results demonstrate D3's superior computational efficiency. Unlike deep learning-based classifiers, D3 is training-free, eliminating substantial preprocessing and training overheads. For instance, on our training set (192,000 samples), Demamba requires 10.45 hours per epoch (batch size 1) and 3.25 hours per epoch (batch size 32). When the training set is set to a larger open-source dataset (e.g., GenVideo training set with a total of 2,262,086 samples), the advantages of D3 become even more pronounced.

In addition, D3 maintains optimal efficiency during inference (56s per 1,000 samples using XCLIP-B/16). D3 further supports lightweight networks (e.g., 40s per 1,000 samples via MobileNet-v3), enabling enhanced computational efficiency and localized deployment.


\begin{figure}[!tp]
    \centering
    \includegraphics[width=\columnwidth]{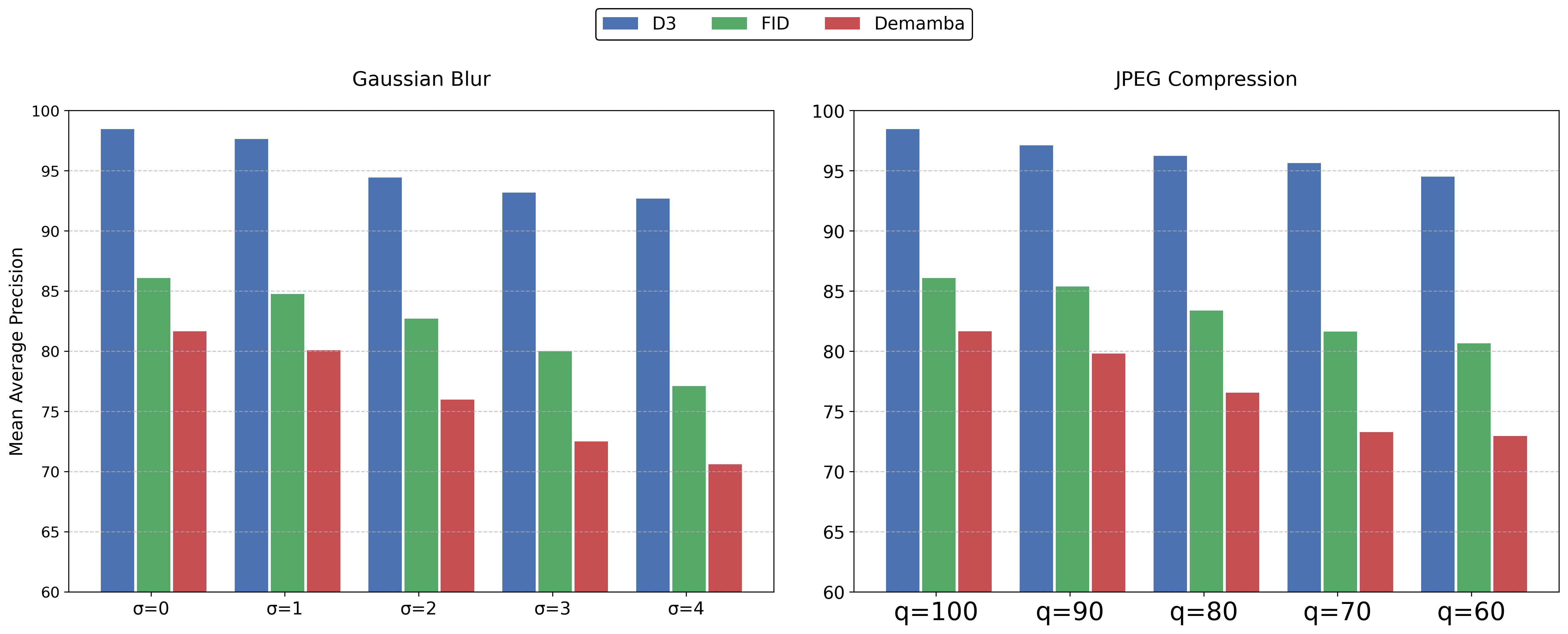}
    \caption{Detection results (mAP) of baselines and D3 against post-processing operations on Genvideo.}
    \label{fig:rob2}
\end{figure}

\begin{figure}[!th]
    \centering
    \includegraphics[width=0.45\textwidth]{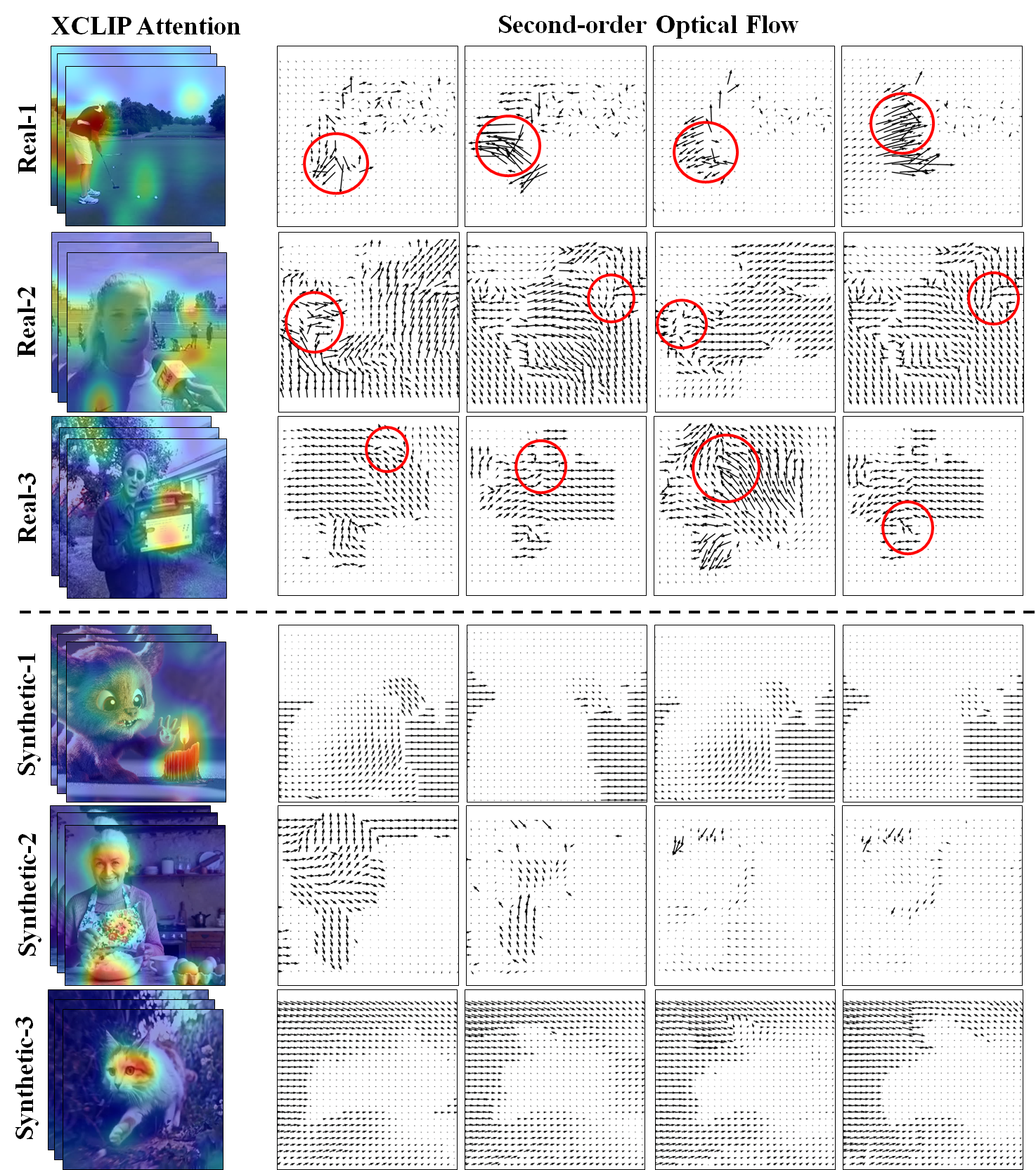}
        \caption{Visualizations of XCLIP attention and 2nd-order flows.}
    \label{fig:heatmap}
\end{figure}

\begin{table*}[!t]
    \centering
    \begin{tabular}{ccccccccccccccc}
        \toprule
        \multirow{2}{*}[-0.5ex]{Visual Encoder} & \multicolumn{5}{c}{Gaussian Blur}  & \multicolumn{5}{c}{JEPG Compression} \\ 
        \cmidrule(r){2-6}
        \cmidrule(r){7-11}
        ~ & $\sigma = 0$ & $\sigma = 1$ & $\sigma = 2$ & $\sigma = 3$ & $\sigma = 4$  & $q = 100$ & $q = 90$ & $q = 80$ & $q = 70$ & $q = 60$
        \\ 
        \midrule
        DINOv2-B & 96.84  & 96.00  & 93.39  & 90.42  & 88.34  & 96.84  & 95.60  & 94.93  & 94.36  & 93.59  \\ 
        DINOv2-L & 96.23  & 95.28  & 92.00  & 88.85  & 87.10  & 96.23  & 94.65  & 93.93  & 93.16  & 92.53  \\ 
        CLIP-B/16 & 97.82  & 97.16  & 83.70  & 83.20  & 84.89  & 97.82  & 84.86  & 83.31  & 81.59  & 78.81  \\ 
       XCLIP-B/16 & \textbf{98.46}  & \underline{97.63}  & \underline{94.43}  & \underline{93.19}  & \underline{92.69}  & \textbf{98.46}  & \underline{97.11}  & \underline{96.24}  & \underline{95.63}  & \underline{94.50}  \\ 
        CLIP-B/32 & 97.71  & 97.05  & 84.50  & 85.13  & 86.56  & 97.71  & 91.65  & 89.79  & 88.04  & 86.40  \\ 
       XCLIP-B/32 & \underline{98.15}  & \textbf{97.64}  & \textbf{95.51}  & \textbf{94.22}  & \textbf{93.59}  & \underline{98.15}  & \textbf{97.59}  & \textbf{97.37}  & \textbf{97.25}  & \textbf{96.93}  \\ 
        ResNet18 & 97.35  & 96.60  & 92.84  & 90.70  & 90.03  & 97.35  & 96.61  & 95.48  & 94.09  & 92.48  \\ 
        VGG16 & 97.21  & 95.56  & 90.91  & 87.28  & 85.60  & 97.21  & 94.06  & 92.35  & 89.47  & 87.58 \\
        EfficientNet-b4 & 96.53  & 95.44  & 90.63  & 88.59  & 88.13  & 96.53  & 94.81  & 93.37  & 91.85  & 90.54  \\ 
        MobileNet-v3 & 96.86  & 95.92  & 87.91  & 84.66  & 84.38  & 96.86  & 94.58  & 91.29  & 88.31  & 85.95 \\ 
        \bottomrule
    \end{tabular}
    \caption{Detection results of D3 against post-processing operations on GenVideo.}
    \label{tab:robust}
\end{table*}

\subsection{Qualitative Analysis}
To further demonstrate the effectiveness of D3, we conduct qualitative analysis by visualizing XCLIP attention and second-order optical flow, as shown in Figure~\ref{fig:heatmap}.

We can see from Figure~\ref{fig:heatmap} that variations in 2nd-order flow appear around moving semantics or objects (highlighted by the XCLIP attention). This is aligned with physical principles (e.g., Newtonian inertia) stating that the object's motion in real-world scenarios follows high-order dynamics.

Besides, the results demonstrate that current video generators fail to learn similar second-order patterns from real videos well (see second-order optical flow in Figure~\ref{fig:heatmap}), which explains the effective detection performance of D3 using second-order video features.

\subsection{Robustness to Post-processing Operations}

In real-world scenarios, videos are seldom pristine. As videos circulate in cyberspace, they are continuously subjected to compression and interference, potentially leading to a performance degradation of video detectors.
To address this, in this section, we evaluate the robustness of D3 against various post-processing operations. Specifically, we consider five different levels of Gaussian blur ($\sigma$ = 0, 1, 2, 3, 4) and JPEG compression with five different quality factors ($q$ = 100, 90, 80, 70, 60). We use the mean AP on GenVideo as the metric.

Figure~\ref{fig:rob2} reveals detection results of baselines and D3 (using XCLIP-B/16 and L2 Distance) against post-processing, with D3 demonstrating the strongest robustness through minimal degradation under Gaussian blur and JPEG compression. As the degree of post-processing operations increases, D3 maintains high detection performance, indicating its excellent robustness against post-processing operations. However, FID and Demamba show pronounced vulnerability, suffering significant performance declines, reflecting their reliance on high-frequency details and susceptibility to spectral artifacts. 

Table~\ref{tab:robust} reports the results of the robustness experiments of different variants of D3. As can be seen, different visual encoders exhibit varying levels of robustness. For example, MobileNet-V3 shows a noticeable performance drop when confronted with ($\sigma$ = 4) Gaussian blur (from 96.86\% to 84.38\%), whereas XCLIP-B/16 exhibits a smaller decrease (from 98.46\% to 92.69\%). This is understandable because MobileNet is a lightweight vision network with a smaller parameter scale. These results reveal that the detection robustness using second-order features depends on the stability of the feature space. Therefore when using self-supervised visual models based on ViT (e.g., XCLIP-B/16), D3 demonstrates better robustness.

\begin{table*}[!t]
    \centering
    \begin{tabular}{lcccccccc}
    \toprule
    \multirow{2}{*}[-0.5ex]{Visual Encoder} & \multicolumn{2}{c}{GenVideo} & \multicolumn{2}{c}{EvalCrafter} & \multicolumn{2}{c}{VideoPhy} & \multicolumn{2}{c}{VidProM} \\ 
    \cmidrule(r){2-3} \cmidrule(r){4-5} \cmidrule(r){6-7} \cmidrule(r){8-9}
    ~ & L2 & Cos & L2 & Cos & L2 & Cos & L2 & Cos \\
    \midrule
    DINOv2-B & 95.84  & 87.17  & 96.76  & 89.31  & 93.98  & 82.14  & 82.17  & 73.23  \\ 
    DINOv2-L & 94.92  & 85.33  & 95.84  & 87.31  & 92.49  & 79.12  & 80.90  & 70.83  \\ 
    CLIP-B/16 & \underline{97.00}  & 87.82  & 97.63  & 89.82  & 97.01  & 86.24  & 84.79  & 75.77  \\ 
   XCLIP-B/16 & \textbf{97.72}  & \underline{91.30}  & \textbf{98.24}  & \underline{92.81}  & \underline{97.14}  & \underline{89.10}  & \textbf{87.08}  & \textbf{79.87}  \\ 
    CLIP-B/32 & 96.73  & 87.87  & 97.26  & 89.53  & 96.61  & 87.04  & 83.97  & 75.52  \\ 
   XCLIP-B/32 & 96.99  & 90.43  & \underline{97.72}  & 92.31  & 96.35  & 88.74  & \underline{85.57}  & \underline{79.62}  \\ 
    ResNet-18 & 96.39  & 89.73  & 97.26  & 91.64  & 95.67  & 86.83  & 81.59  & 75.68  \\ 
      VGG-16 & 96.97  & \textbf{92.63}  & \underline{97.84}  & \textbf{94.16}  & \textbf{97.50}  & \textbf{91.21}  & 81.54  & 77.02  \\ 
EfficientNet-B4 & 94.28  & 85.51  & 95.49  & 88.08  & 92.46  & 82.40  & 80.73  & 73.00  \\ 
    MobileNet-V3 & 95.47  & 87.14  & 96.48  & 89.50  & 94.70  & 84.71  & 80.76  & 73.74 \\ 
    \bottomrule
    \end{tabular}
    \caption{
    Ablation studies of visual encoder backbones and the type of first-order features (L2 Distance or Cosine Similarity).
    }
    \label{tab:ablation}
\end{table*}


\section{Ablation Studies}
\subsection{Impact of Visual Encoders}\label{Effect of Visual Encoders}

So far, we have demonstrated the effectiveness of D3 in generalized detection. In the above experiments, D3 utilizes the pre-trained XCLIP-ViT-B/16 model, which raises a new question of whether generated video detection using second-order coefficients relies on large-scale visual encoders. To address this, we conducted several ablation experiments using different visual encoders. We adopt several large-scale self-supervised models based on ViT, including CLIP-ViT, XCLIP-ViT, DINOv2, and their variants with different patch sizes or parameter scales. Additionally, we adopt CNN-based models for classification, e.g., ResNet18, VGG16, EfficientNet-B4, or lightweight networks like MobileNet. These CNN-based models are all pre-trained on ImageNet.

The results of the ablation experiments are shown in Table~\ref{tab:ablation}. An intuitive conclusion is that large-scale visual encoders perform best in our experiments, e.g., CLIP-ViT-B/16 and XCLIP-ViT-B/16. Among the ViT-based models, the impact of patch size and parameter scale is negligible, as evidenced by the minimal differences between XCLIP-ViT-B/16 and XCLIP-ViT-B/32. 

Nonetheless, despite the significant differences in parameter scale or model architecture among these visual encoders, the differences in detection performance are small. For example, as a lightweight model, MobileNet-V3 still produces an impressive result of 96.31\%. This phenomenon is encouraging because it confirms that second-order features remain meaningful across different encoder feature spaces. This insight helps us understand the shortcomings of current video generators in simulating reality.

\subsection{Impact of First-order Features}

In this section, we delve into the impact of different first-order feature extraction methods. Specifically, we adopt L2 distance and cosine similarity separately as the first-order feature. The experimental results are shown in Table~\ref{tab:ablation}.

These results indicate that using L2 distance as the first-order feature yields better performance. The key takeaway is that L2 distance evaluates the absolute distance between inter-frame features, while cosine similarity evaluates the relative distance. Cosine similarity can better mitigate the effects of differing feature dimensions. However, in our experiments, the visual encoder is fixed, and therefore, the output feature dimensions are fixed. Besides, cosine similarity will be influenced by the features of the initial frame, whereas L2 distance can accurately reflect the extent of video change within this fixed feature space. Therefore, it provides more effective information.

\begin{table}[!t]
    \centering
    \resizebox{\columnwidth}{!}{
        \begin{tabular}{l c c c c c}
            \toprule
            Detection & \multicolumn{2}{c}{GenVideo} & \multicolumn{2}{c}{EvalCrafter}\\
            \cmidrule(r){2-3} \cmidrule(r){4-5}
            Method & mAP$\uparrow$ & Avg. AUC$\uparrow$ & mAP$\uparrow$ & Avg. AUC$\uparrow$ \\ 
            \midrule
            D3 (1st-Order) & 95.69 & 93.45 & 86.40 & 85.17 \\ 
            D3 (2nd-Order) & \textbf{98.46}  & \textbf{97.72}  & \textbf{98.87}  & \textbf{98.24}  \\ 
            \midrule
            ~ & \multicolumn{2}{c}{VideoPhy} & \multicolumn{2}{c}{VidProM}\\
            \cmidrule(r){2-3} \cmidrule(r){4-5}
            ~ & mAP$\uparrow$ & Avg. AUC$\uparrow$ & mAP$\uparrow$ & Avg. AUC$\uparrow$ \\ 
            \midrule
            D3 (1st-Order) & 86.06 & 84.22 & 80.61 & 77.31 \\ 
            D3 (2nd-Order) & \textbf{99.16}  & \textbf{97.14}  & \textbf{88.46}  & \textbf{87.08} \\ 
            \bottomrule
        \end{tabular}
    }
    \caption{Ablation studies of feature order on 4 datasets.}
    \label{tab:order}
\end{table}

\subsection{Second-Order vs. First-Order Features}\label{Second-Order vs. First-Order Features}

In this section, we explore the impact of feature order on the D3 method. Specifically, we replace the second-order feature standard deviation in the original scheme with the first-order feature standard deviation and evaluate it on 4 datasets. Table~\ref{tab:order} presents the results of our ablation study.
As shown, D3 using first-order features achieves good performance (95.69\% mAP and 93.45\% Avg. AUC) on GenVideo. However, this performance cannot be generalized to more challenging datasets. 

Overall, using second-order features provides stronger detection capability. These results suggest that, although there are some differences in the first-order features between real and AI-generated videos (as shown in Figure~\ref{fig:flow}), such differences are not universal, while second-order differential features are more powerful.

\section{Conclusion and Outlook}\label{Conclusion}
This paper bridges the theory of second-order control systems from Newtonian mechanics with video analysis by extending second-order central difference features for temporal artifact detection. Our systematic analysis reveals that existing AI-generated videos diverge from real videos in their second-order feature, establishing a novel physical perspective for investigating temporal artifacts in synthetic content. Building on these insights, we proposed D3, an innovative, training-free AI-generated video detection framework. By measuring the volatility of second-order features in videos through standard deviation, D3 achieves generalizable detection of AI-generated videos. 
Through extensive experiments, we demonstrate that D3 achieves state-of-the-art performance in detecting AI-generated videos across various generative models as well as strong robustness against post-processing operations.

Our approach paves a new way for understanding and differentiating between real and AI-generated videos based on the connection between video content and fundamental physical principles.  
Future investigations can explore this paradigm by examining additional dimensions (e.g., temporal channel relationships, RGB color space distributions, or bitrate characteristics) to deepen our understanding of generation artifacts. 
We believe this paper will inspire further research into the artifacts of AI-generated videos and contribute to the development of generalized detection.

\bibliographystyle{ieeenat_fullname}
\bibliography{main}

\end{document}